\documentclass{article}

\usepackage{tccml_iclr2025_conference, times}

\usepackage[utf8]{inputenc} 
\usepackage[T1]{fontenc}    
\usepackage{hyperref}       
\usepackage{url}            
\usepackage{booktabs}       
\usepackage{amsfonts}       
\usepackage{nicefrac}       
\usepackage{microtype}      
\usepackage{multirow}       
\usepackage{graphicx}       
\usepackage{subcaption}     

\setcitestyle{numbers}

\title{Improving Tropical Cyclone Forecasting With Video Diffusion Models}
\author{%
  Zhibo Ren \\
  Department of Computing \\
  Imperial College London \\ 
  \texttt{zhibo.ren23@imperial.ac.uk}
  \And
  Pritthijit Nath \\
  Department of Applied Math and Theoretical Physics \\
  University of Cambridge \\  \texttt{pn341@cam.ac.uk}
  \And
  Pancham Shukla \\
  Department of Computing \\
  Imperial College London \\   \texttt{panchamkumar.shukla@imperial.ac.uk}
}

\iclrfinalcopy

\begin{document}
\setcitestyle{square}
\maketitle

\begin{abstract}
Tropical cyclone (TC) forecasting is crucial for disaster preparedness and mitigation. While recent deep learning approaches have shown promise, existing methods often treat TC evolution as a series of independent frame-to-frame predictions, limiting their ability to capture long-term dynamics. We present a novel application of video diffusion models for TC forecasting that explicitly models temporal dependencies through additional temporal layers. Our approach enables the model to generate multiple frames simultaneously, better capturing cyclone evolution patterns. We introduce a two-stage training strategy that significantly improves individual-frame quality and performance in low-data regimes. Experimental results show our method outperforms the previous approach of Nath et al. by 19.3\% in MAE, 16.2\% in PSNR, and 36.1\% in SSIM. Most notably, we extend the reliable forecasting horizon from 36 to 50 hours. Through comprehensive evaluation using both traditional metrics and Fréchet Video Distance (FVD), we demonstrate that our approach produces more temporally coherent forecasts while maintaining competitive single-frame quality. Code accessible at %
\url{https://github.com/Ren-creater/forecast-video-diffmodels}. 
\end{abstract}

\section{Introduction}
Climate change is a pressing global issue causing unprecedented changes in the Earth's climate system, with tropical cyclones (TCs) receiving particular attention due to their increasing intensity and devastating impacts~\cite{knutson_tropical_2020}. These extreme weather events pose a growing threat to global communities, causing extensive human suffering and economic losses~\cite{krichene_social_2023}. Recent studies indicate that warming oceans (due to climate change) are leading to more powerful storms~\cite{emanuel_response_2021}, making accurate TC forecasting increasingly critical for effective emergency preparedness and disaster response. 

Traditional numerical weather prediction (NWP) methods~\cite{schalkwijk_weather_2015}, which rely on solving complex physical equations, have been the foundation of TC forecasting for decades. While these methods are rooted in well-established meteorological principles, they are computationally demanding and can encounter difficulties in capturing the highly non-linear and chaotic nature of atmospheric dynamics~\cite{huang_tcp-diffusion:_2024}. Such limitations, coupled with the increasing demand for more precise and timely predictions whilst being computationally inexpensive, have spurred interest in alternative or complementary approaches.

Machine learning (ML) has emerged as a promising tool for addressing these challenges, offering the potential to enhance or even replace traditional NWP methods. ML models can leverage vast amounts of observational and simulation data to identify intricate patterns and provide rapid inference~\cite{pouyanfar_survey_2019}. Among these, deep learning (DL) techniques have been particularly effective, demonstrating success in a variety of TC-related tasks, including cyclone detection~\cite{accarino_ensemble_2023}, track prediction~\cite{kovordanyi_cyclone_2009}, and precipitation forecasting~\cite{leinonen_latent_2023}. Recent work by Nath et al.~\cite{nath_forecasting_2024} has further highlighted the potential of diffusion models in TC forecasting. However, their approach generates predictions on a frame-by-frame basis, which may limit its ability to capture critical temporal dependencies that span multiple time steps.

Recent advancements in video diffusion models have opened new possibilities for tackling such limitations. These models, which have achieved remarkable performance in areas such as video synthesis and editing~\cite{ho_video_2022, blattmann_stable_2023}, explicitly account for temporal dynamics through specialised temporal layers. By generating multiple frames simultaneously, video diffusion models ensure better temporal coherence, a feature that is particularly crucial for weather forecasting applications where the evolution of atmospheric systems over time must be accurately represented.

Despite the recent progress, the challenge of TC forecasting still remains formidable. Skilful prediction requires models that can simultaneously capture spatial and temporal dependencies while ensuring stability and accuracy over longer prediction horizons. Although existing approaches have made considerable progress in short-term forecasting, issues such as temporal inconsistency and forecast degradation over time persist, leaving room for further innovation. Addressing these challenges is essential for pushing the boundaries of TC forecasting and delivering solutions that are both robust and practical.

To advance progress in this domain, our work makes the following contributions:

\begin{enumerate}
    \item We present a novel application of video diffusion models for TC forecasting, enabling simultaneous generation of multiple frames to better capture temporal dynamics.
    \item We introduce a two-stage training strategy that significantly improves model training stability and individual frame-quality, especially in low-data regimes.
    \item We demonstrate superior long-horizon forecasting capability, extending reliable predictions from 36 (as demonstrated by Nath et al.) to 50 hours.
    \item We establish Fréchet Video Distance (FVD)~\cite{unterthiner_towards_2019} as a more suitable metric for evaluating TC forecasts, providing better assessment of temporal coherence.
\end{enumerate}

\section{Method}
Our approach builds upon the work of Nath et al.~\cite{nath_forecasting_2024}, extending their diffusion-based framework to incorporate temporal dynamics through video generation. The key insight is that treating TC evolution as a continuous process, rather than as a series of independent frame predictions, allows more effective modelling of storm dynamics.

\subsection{Data Processing}
We use the same dataset as Nath et al.~\cite{nath_forecasting_2024}, reorganising it into sequences of 10 consecutive frames for both infrared (IR) satellite images and corresponding ERA5 data. This results in 1,092 video sequences for training and 335 for testing. To handle corrupted data points, \texttt{NaN} values are replaced with zeros, and a consistent mask is applied during generation to ensure they do not influence the output.

\subsection{Video Diffusion Model}
We adopt a 3D UNet architecture (illustrated in Figure~\ref{fig:cddpm_flow}) following a design similar to that proposed by Ho et al.~\cite{ho_video_2022}. The model utilises temporal convolutions and attention mechanisms to generate sequences of 64x64 IR 10.8µm satellite imagery, conditioned on both the initial IR frame and corresponding ERA5 meteorological data. In particular, the model generates 10 forecast frames simultaneously, enabling it to better capture the temporal evolution of cyclonic systems. This architecture incorporates both classifier-free guidance~\cite{ho_classifier-free_2022} and dynamic thresholding for maintaining output quality within the normalised range. Detailed hyperparameter configurations used in model training are provided in Appendix~\ref{tab:hyperparameters}.

\subsection{Two-Stage Training}


We introduce a two-stage training strategy crucial for model performance. In the low-data regime, we first conduct a single-frame stage of 200 epochs, focusing on individual frame prediction to establish spatial understanding. Then, in the multi-frame stage, we train on 10-frame sequences for an additional 200 epochs to learn temporal dynamics. For the full dataset, we perform a single-frame stage of 100 epochs, followed by a multi-frame stage of 300 epochs to better capture temporal dynamics.

This curriculum learning approach significantly improves training stability and model performance. Notably, in the low-data regime experiments (see Section \ref{sec:two_stage_training}), while training without the first stage achieves similar FVD scores (402.98 v/s 402.03 w/ the first stage), the two-stage approach achieves superior single-frame quality with FID of 0.49 compared to 1.26 for direct training without the first stage. This gain in performance demonstrates that the initial single-frame training stage is essential for preserving frame-level accuracy even as we optimise for temporal coherence.

In the full dataset scenario, we found that adjusting the epoch distribution (100/300 v/s 200/200) provides better overall results while maintaining the crucial single-frame quality benefits of two-stage training. This suggests that when sufficient data are available, allocating more training time to multi-frame prediction is beneficial, as long as the initial single-frame training stage is preserved.

\section{Experiments}

We evaluated our model using multiple metrics to assess both single-frame quality and temporal coherence. Besides traditional metrics (MAE, PSNR, SSIM, FID) used in previous work, we introduce FVD~\cite{unterthiner_towards_2019} to better evaluate temporal consistency. All experiments used a learning rate of $3 \times 10^{-4}$ and are performed on a single NVIDIA L40 GPU, demonstrating remarkable computational efficiency of our approach (in contrast to conventional methods).

\subsection{Data Efficiency Study}
\label{sec:two_stage_training}
To investigate model performance in low-data regimes, we trained both models using only cyclones in the North Indian Ocean basin. The results shown in Table \ref{tab:low-data} demonstrate our model's superior data efficiency:

\begin{table}[h]
\centering
\caption{Performance comparison in low-data regime}
\label{tab:low-data}
\begin{tabular}{@{}lccccc@{}}
\toprule
Method & MAE $\downarrow$ & PSNR $\uparrow$ & SSIM $\uparrow$ & FID $\downarrow$ & FVD $\downarrow$ \\
\midrule
Baseline (Nath et al.~\cite{nath_forecasting_2024}) & 0.2846 & 18.07 & 0.4353 & \textbf{0.3298} & 706.11 \\
Video Diffusion (w/o two-stage) & 0.2647 & \textbf{20.72} & \textbf{0.6522} & 1.2633 & 402.98 \\
Video Diffusion (with two-stage) & \textbf{0.2300} & 20.62 & 0.6387 & 0.4955 & \textbf{402.03} \\
\bottomrule
\end{tabular}
\end{table}

\textbf{Impact of Two-Stage Training}:
The two-stage training strategy proves crucial for maintaining single-frame quality while improving temporal coherence. Without two-stage training, the model achieves similar FVD scores but suffers in single-frame quality (FID increases from 0.49 to 1.26). 

\subsection{Results and Discussion}

\begin{table}[h]
\centering
\caption{Comparison with baseline model on 10-frame prediction task}
\label{tab:comparison}
\begin{tabular}{@{}lccccc@{}}
\toprule
Method & MAE $\downarrow$ & PSNR $\uparrow$ & SSIM $\uparrow$ & FID $\downarrow$ & FVD $\downarrow$ \\
\midrule
Baseline (Nath et al.~\cite{nath_forecasting_2024}) & 0.2209 & 22.49 & 0.5235 & \textbf{0.2288} & 445.83 \\
Video Diffusion & \textbf{0.1781} & \textbf{26.13} & \textbf{0.7123} & 0.2344 & \textbf{242.41} \\
Improvement & 19.3\% & 16.2\% & 36.1\% & -2.4\% & 45.6\% \\
\bottomrule
\end{tabular}
\end{table}

Our model significantly outperforms the baseline (Nath et al.~\cite{nath_forecasting_2024}) across most metrics, with particularly strong improvements in temporal coherence as measured by FVD (45.6\% reduction). While maintaining similar single-frame quality (FID), our model achieves substantial improvements in MAE (19.3\%), PSNR (16.2\%), as well as SSIM (36.1\%).

\begin{figure}[h]
\centering
\includegraphics[width=12cm]{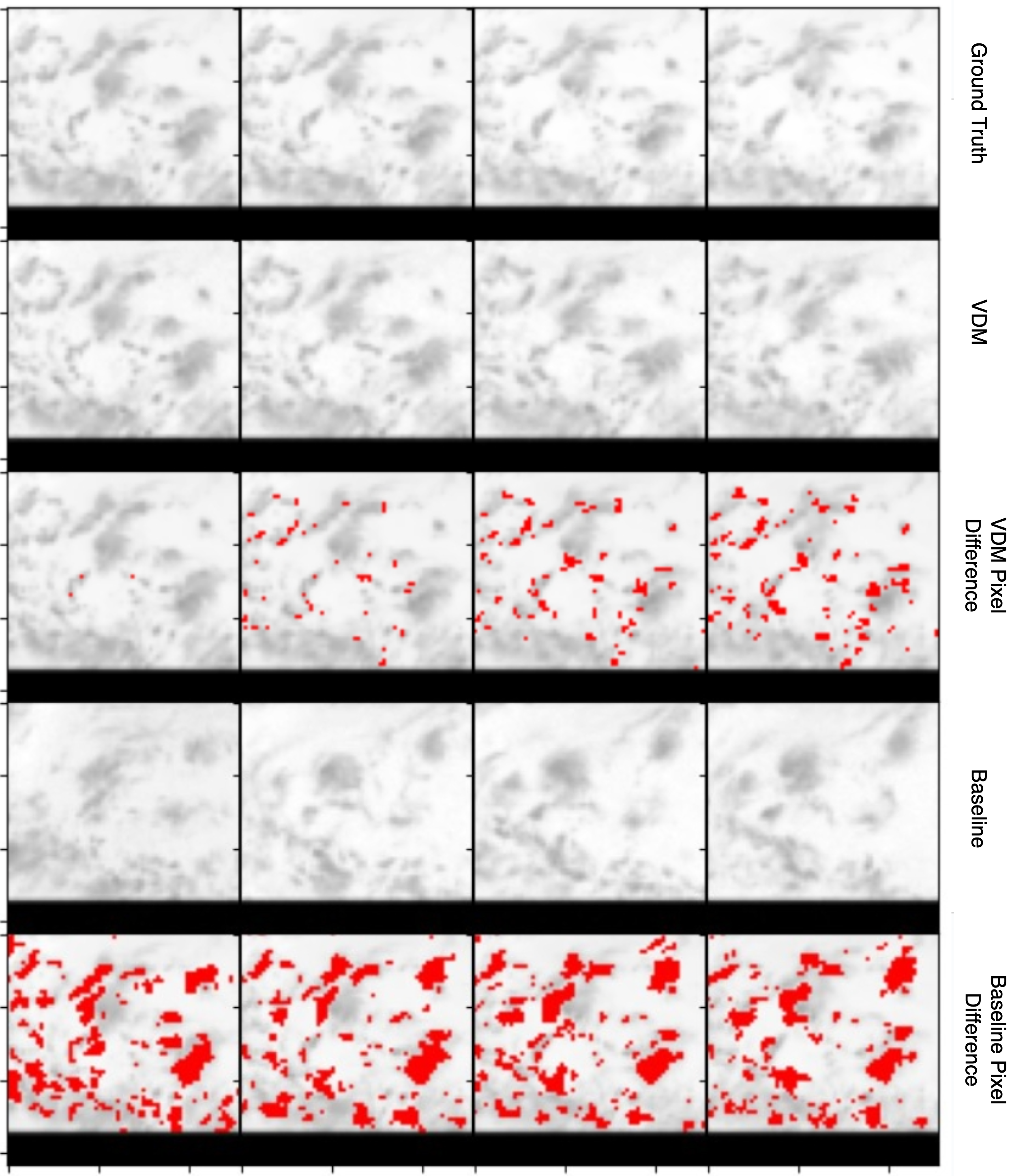}
\caption{Qualitative comparison of TC forecasting results on the first four frames generated. From top to bottom: (1) Ground truth, (2) our VDM predictions, (3) the difference between VDM prediction and ground truth, (4) Nath et al.'s predictions, and (5) the difference between Nath et al.'s predictions and ground truth. Our VDM method demonstrates improved temporal consistency and more accurate TC evolution patterns.}
\label{fig:comparison}
\end{figure}

Visual inspection (see Figure \ref{fig:comparison}) of the samples reveals that our model handles complex cloud patterns more effectively than the baseline. While 
the baseline often struggles with dense cloud regions, our approach maintains better cloud cover consistency with the ground truth. This improvement can be greatly attributed to the model's improved ability to consider temporal context during generation.

\subsection{Long-horizon Forecasting}
Following Nath et al.~\cite{nath_forecasting_2024}, we performed similar experiments to forecast the entire duration of all cyclones in the testing split, organised region-wise. Nath et al.~\cite{nath_forecasting_2024} identified a reliable forecasting horizon of 36 hours, beyond which sharp declines in forecast accuracy, as indicated by SSIM charts, are observed. Integrating our model into their cascaded pipeline extends this reliable horizon from 36 to 50 hours, as shown by SSIM charts (see charts in \ref{sec:appendixC}). Furthermore, the minimum SSIM values predicted by our model consistently remain higher than those of the baseline, suggesting a better preservation of cyclone structure over extended periods.

\section{Conclusion}
We present a novel application of video diffusion models for forecasting TCs that significantly improves both temporal coherence as well as forecast horizon. Our two-stage training strategy successfully addresses the challenge of maintaining single-frame quality while optimising for temporal dynamics. Experimental results demonstrate substantial improvements across multiple metrics, with particularly strong gains in temporal consistency (45.6\% reduction in FVD) while preserving single-frame fidelity with a competitive FID of 0.23. Most notably, our method extends the reliable forecasting horizon from 36 to 50 hours, representing a significant step forward in enhancing long-term TC prediction capabilities.

For future work, several promising directions emerge, including integration of multiple satellite channels (beyond IR bands), increasing the number of frames generated in a single forward pass beyond the current 10-frame limit, investigation of physics-informed loss functions to better preserve cyclone dynamics (ensuring physical consistency), and using VDMs as extension to other crucial weather forecasting applications in need of improvements in temporal coherence capabilities.

\section*{Acknowledgements and Disclosure of Funding}
P. Nath was supported by the \href{https://ai4er-cdt.esc.cam.ac.uk/}{UKRI Centre for Doctoral Training in Application of Artificial Intelligence to the study of Environmental Risks} [EP/S022961/1]. 

\clearpage
\bibliographystyle{vancouver}
\bibliography{bibliography}

\clearpage

\appendix
\renewcommand\thesection{Appendix \Alph{section}}
\renewcommand\thesubsection{\Alph{section}.\arabic{subsection}} 
\renewcommand\thetable{\Alph{section}.\arabic{table}}  
\renewcommand\thefigure{\Alph{section}.\arabic{figure}}  
\setcounter{table}{0}
\setcounter{figure}{0}

\section{Model Architecture}
\label{sec:appendixA}
\begin{figure}[!h]
\centering
\includegraphics[width=\textwidth]{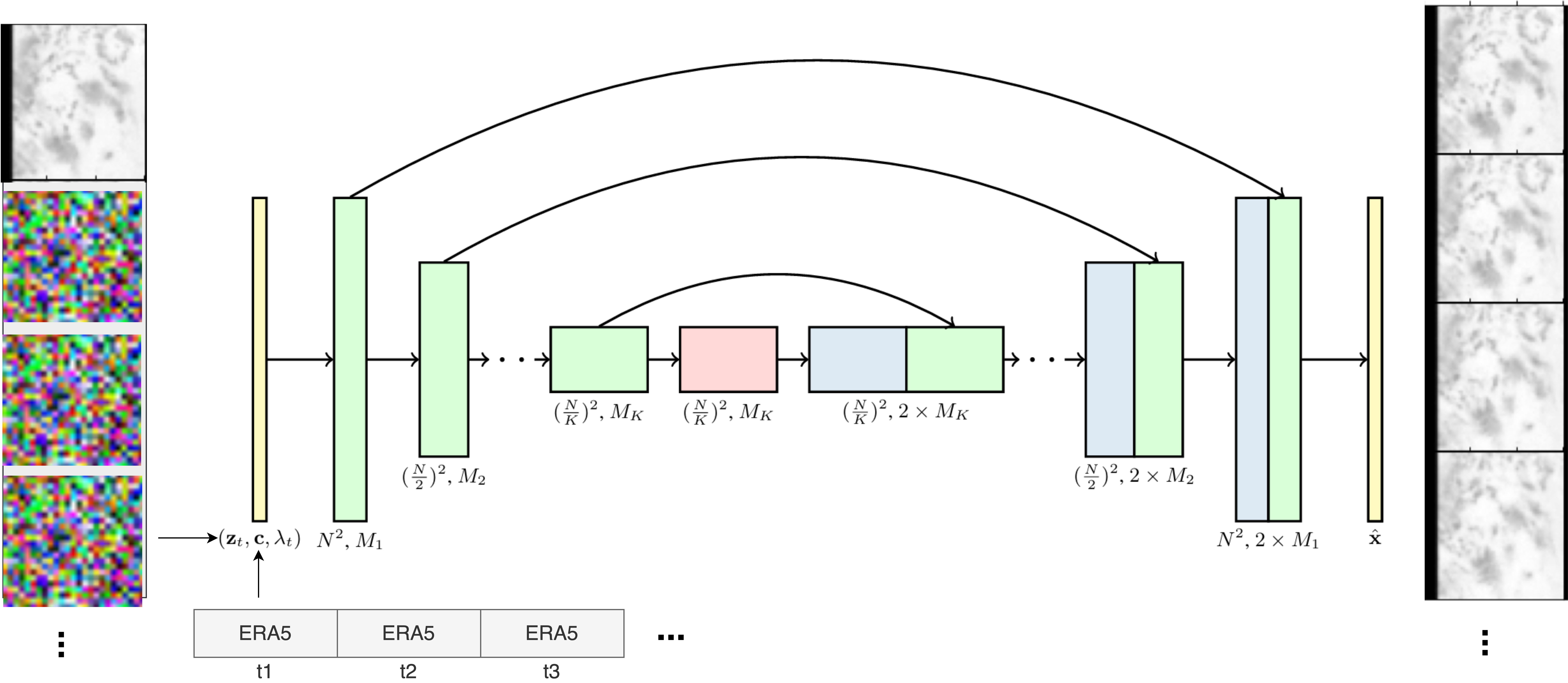}
\caption{Illustration of the model pipeline. Our VDM model takes as input a noisy image \( \mathbf{z}_t \), conditioning variables \( \mathbf{c} \), and a timestep embedding \( \lambda_t \), and progressively denoises the sample using a U-Net-based architecture with skip connections. ERA5 data from multiple timesteps (\( t_1, t_2, t_3 \)) is used as conditioning information. Output \( \hat{\mathbf{x}} \) represents the denoised prediction.}
\label{fig:cddpm_flow}
\end{figure}

\section{Implementation Details}
\label{sec:appendixB}
\begin{table}[!h]
\centering
\caption{Hyperparameters used in model training}
\label{tab:hyperparameters}
\begin{tabular}{lcc}
\toprule
\textbf{Hyperparameter} & \textbf{Global Dataset} & \textbf{North Indian Ocean Region} \\
\midrule
Batch Size & 1 & 1 \\
Sequence Length & 10 & 10 \\
Learning Rate & $3 \times 10^{-4}$ & $3 \times 10^{-4}$ \\
Guidance Scale & 3.0 & 3.0 \\
Epochs Stage 1 (Single-frame) & 100 & 200 \\
Epochs Stage 2 (Multi-frame) & 300 & 200 \\
\bottomrule
\end{tabular}
\end{table}

\clearpage

\section{Additional Results}
\label{sec:appendixC}
\subsection{Forecast SSIM}
\begin{figure}[!h]
    \centering
    \begin{tabular}{@{}ccc@{}}
        \begin{minipage}{0.3\textwidth}
            \centering
            \includegraphics[width=\textwidth]{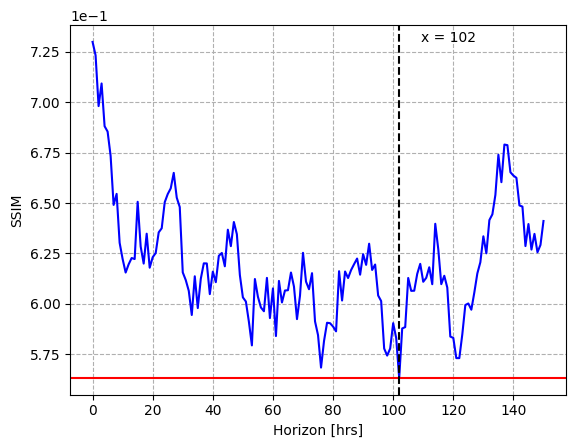} \\
            (a) Mocha \\
            (North Indian Ocean)
        \end{minipage} &
        \begin{minipage}{0.3\textwidth}
            \centering
            \includegraphics[width=\textwidth]{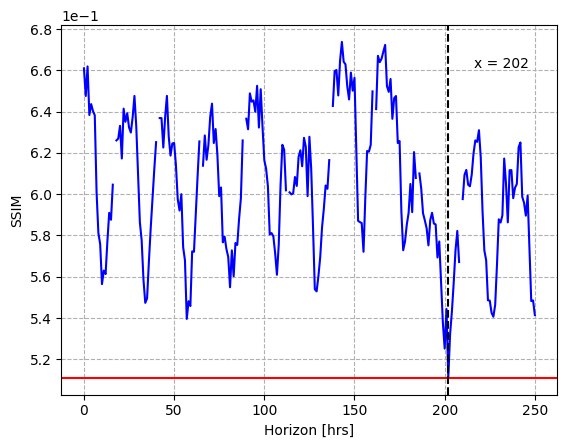} \\
            (b) Ida \\
            (North Atlantic Ocean)
        \end{minipage} &
        \begin{minipage}{0.3\textwidth}
            \centering
            \includegraphics[width=\textwidth]{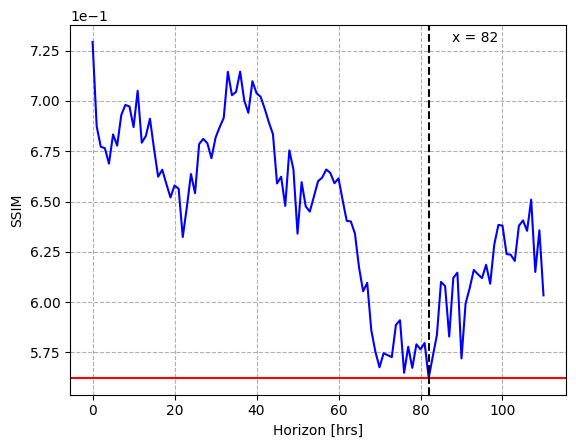} \\
            (c) Roslyn \\
            (Eastern Pacific Ocean)
        \end{minipage} \\
        \begin{minipage}{0.3\textwidth}
            \centering
            \includegraphics[width=\textwidth]{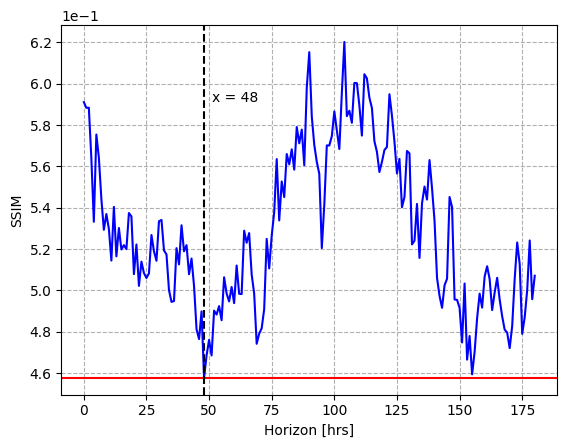} \\
            (d) Molave \\
            (Western Pacific Ocean)
        \end{minipage} &
        \begin{minipage}{0.3\textwidth}
            \centering
            \includegraphics[width=\textwidth]{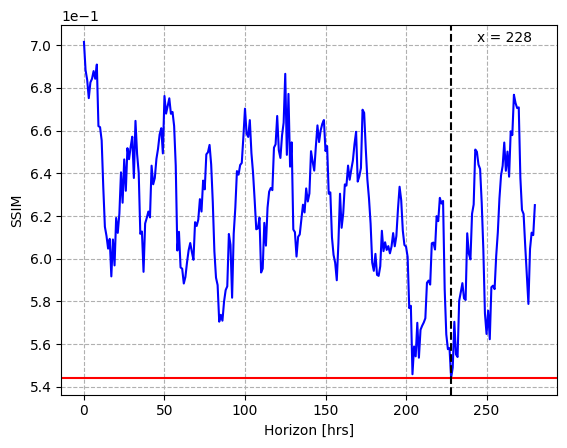} \\
            (e) Gombe \\
            (SW Indian Ocean)
        \end{minipage} &
        \begin{minipage}{0.3\textwidth}
            \centering
            \includegraphics[width=\textwidth]{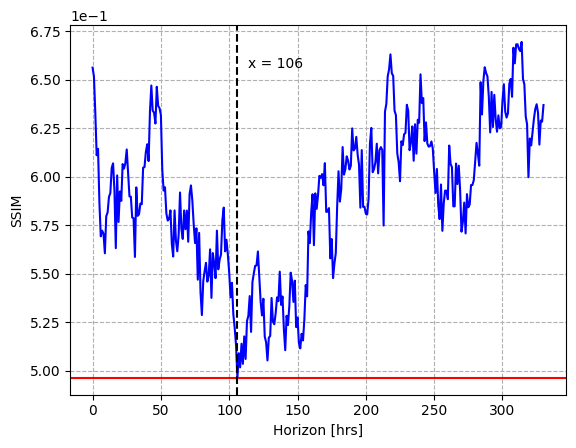} \\
            (f) Veronica \\
            (Australia)
        \end{minipage}
    \end{tabular}
    \caption{SSIM values over the entire cyclonic duration. The dashed lines indicate the hourly marks at which the minimum SSIM values are obtained for each cyclone.}
    \label{fig:forecast_ssim}
\end{figure}

\end{document}